\documentclass[letterpaper, 10 pt, conference]{ieeeconf}  
\usepackage{amsmath,amsfonts}
\usepackage{algorithmic}
\usepackage{algorithm}
\usepackage{array}
\usepackage[caption=false,font=normalsize,labelfont=sf,textfont=sf]{subfig}
\usepackage{textcomp}
\usepackage{stfloats}
\usepackage{url}
\usepackage{verbatim}
\usepackage{graphicx}
\usepackage[utf8]{inputenc}
\usepackage{textgreek}
\usepackage{cite}
\usepackage{booktabs}
\usepackage{threeparttable}
\usepackage{makecell}
\usepackage{color}
\usepackage[table]{xcolor}
\usepackage{colortbl}
\usepackage{multicol}
\usepackage{multirow}
\usepackage{soul}
\usepackage{adjustbox}
\usepackage{multicol}
\usepackage{multirow}
\usepackage{booktabs}
\usepackage{threeparttable}
\usepackage{graphicx}
\usepackage{balance}
\usepackage{etoolbox,siunitx}
\usepackage{calc}
\usepackage{pifont,hologo}
\usepackage{nicefrac}
\usepackage{makecell}
\usepackage{svg}
\usepackage{enumerate}
\usepackage{pifont}
\usepackage{bbm}
\usepackage{amsmath, amssymb, bm, booktabs}
\usepackage{censor}
\usepackage{tabularx}

\definecolor{bestcolor}{RGB}{250,220,180}     
\definecolor{secondcolor}{RGB}{255,245,220}   

\newcommand{\bestbox}{%
  \raisebox{0.25ex}{\colorbox{bestcolor}{\rule{0pt}{0.8ex}\hspace{1.5em}}}%
}
\newcommand{\secondbox}{%
  \raisebox{0.25ex}{\colorbox{secondcolor}{\rule{0pt}{0.8ex}\hspace{1.5em}}}%
}

\IEEEoverridecommandlockouts                              

\title{\LARGE \bf
PL-LIT: A LiDAR-Inertial-Thermal SLAM Using Point-Line Features and Thermographic Mapping
}

\author{
    Jiawei Xia$^{1,3}$, 
    Yixiao Feng$^{1,2}$, 
    Yongliang Shi$^{1\dagger}$, 
    Chao Gao$^{1\dagger}$, 
    Renjing Xu$^{2}$, 
    Weining Lu$^{1}$,
    Bin Liang$^{1}$
    \thanks{$^{1}$ Tsinghua University}
    \thanks{$^{2}$ The Hong Kong University of Science and Technology (Guangzhou)}
    \thanks{$^{3}$ University of California, Berkeley}
    \thanks{$^{\dagger}$ Corresponding author}
    \thanks{Sponsored by the Xinchen Qihang Co., Ltd.}
}

\begin{document}

\maketitle
\thispagestyle{empty}
\pagestyle{empty}

\begin{abstract}

Thermal imaging is resilient to adverse conditions—such as intense illumination, low-light operation, and fog—and can therefore mitigate odometry degradation when visible-spectrum imagery becomes unreliable. Nevertheless, most thermal cameras employ automatic gain control (AGC), and thermal images often present low global contrast despite containing informative edge structures. These characteristics undermine brightness constancy and cause conventional optical flow tracking-based odometry pipelines that fundamentally rely on the brightness constancy assumption across consecutive frames.
To address these issues, we propose a general LiDAR–Inertial-Thermal SLAM system that accommodates both visible-light and thermal cameras. PL-LIT combines an online photometric calibration module with a deep neural network for point–line feature extraction, enabling more stable and repeatable thermal tracking. For state estimation, we design a tightly coupled LiDAR–Inertial–Thermal formulation within an Error-State Iterated Kalman Filter (ESIKF). We further introduce a line-feature constraint scheme
ensuring the reliability of geometric constraints across varying thermal appearances. In addition, PL-LIT builds a probabilistic thermal-intensity voxel map, which supports real-time thermal anomaly detection. Extensive experiments demonstrate that PL-LIT exhibits generality and robustness in visible-light environments, achieves state-of-the-art performance on long-range thermal infrared datasets, and provides practical safety inspection functionality based on thermographic mapping.

\end{abstract}

\section{INTRODUCTION}

LiDAR-Inertial-Visual SLAM (LIV-SLAM) \cite{zheng2024fastlivo2,zheng2022fast,lin2022r3live} has become a cornerstone for robotic autonomy, enabling reliable localization and mapping in tasks such as navigation and inspection. Nevertheless, existing tightly coupled LIV-SLAM systems exhibit an inherent vulnerability: the visual subsystem is highly sensitive to abrupt illumination changes. In real-world, for instance, strong backlighting can cause severe overexposure, while nighttime conditions often lead to degraded image quality; both factors can critically undermine the robustness of the visual subsystem and even result in complete tracking failure. Thermal imaging offers a complementary sensing modality that remains effective in many of these visually degraded scenarios, thereby mitigating the limitations of conventional visible-light cameras. 

Thermal infrared cameras, with their characteristic passive sensing of long-wave infrared radiation, naturally fit the dual requirements of resisting illumination variation and performing defect inspection. However, applying thermal imaging directly to high-precision SLAM faces specific hurdles: thermal images suffer from low contrast, high noise, and a lack of color and texture details. Furthermore, AGC applies non-linear drastic radiometric rescaling to the raw signal based on scene temperature extremes. This severely disrupts photometric consistency and leads to a significant distribution shift of feature descriptors between consecutive frames, making traditional optical flow or feature-based methods highly prone to failure. Nevertheless, we draw inspiration from the photometric calibration method \cite{das2021online} and the fact that thermal images possess strong structural characteristics—manifested in object edges and linear features—which tend to be more stable than texture \cite{xue2019learning}.

Moreover, current inspection robots typically treat thermal data as independent 2D video streams, causing critical thermal data to lose precise three-dimensional spatial attributes. Consequently, systems cannot automatically associate abnormal heat sources with specific equipment components, often relying on manual secondary verification. This causes the existing detection schemes to not only rely heavily on manual labor. Some thermal imaging temperature measurement methods directly adopt raw data, but this approach is constrained by several issues: extremely low image contrast caused by the absence of AGC and high computational resource requirements, which make it difficult to integrate into an autonomous SLAM system. Meanwhile, we have also noticed that in many cases, detecting the temperature variation is more effective than the temperature threshold.

To address these issues, this paper proposes a tightly-coupled \textbf{P}oint-\textbf{L}ine \textbf{L}iDAR-\textbf{I}nertial-\textbf{T}hermal SLAM system (\textbf{PL-LIT}). The main contributions are as follows:

\begin{itemize}
    \item[1)] System-Level Framework: We propose PL-LIT, which is a tightly-coupled Point-Line LiDAR-Inertial-Thermal SLAM system based on ESIKF that simultaneously performs structural-aware odometry and probabilistic thermal anomaly mapping. PL-LIT also can exhibit generality and robustness in visible-light environments.

    \item[2)] Robust Front-end Design: We develop a structure-aware visual front-end tailored to thermal imaging. Local plane fitting on LiDAR point clouds is first performed to associate reliable depth with thermal feature points. An online photometric calibration module is integrated to mitigate inter-frame nonlinear radiometric fluctuations and improve feature tracking consistency. To address the intrinsic characteristics of thermal imagery—weak texture yet strong structural edges—we employ the PL-Net\cite{xu2024airslam} to extract both point and line features and incorporate structural constraints into the reprojection optimization for pose refinement. This design enhances robustness and accuracy in long-range, illumination-variant urban environment.
    
    \item[3)] Thermal Anomaly Mapping: We propose a thermographic map construction scheme integrated with loop closure detection. This method utilizes thermal images to construct a Probabilistic Consistent Intensity Map, enabling active detection of abnormal temperature variation. Experimental results demonstrate that PL-LIT maintains competitive performance in different environments while innovatively achieving the functionality of identifying thermal anomalies within the map.
\end{itemize}

\section{Related Works}

\subsection{Multi-modal Sensor Fusion SLAM}
Multi-modal sensor fusion has become the standard for robust state estimation in complex environments. Early approaches often adopted a loosely-coupled architecture, treating LiDAR and visual odometry as separate modules. For example, V-LOAM \cite{zhang2014loam} combines a visual odometry front-end with a LiDAR mapping back-end to correct drift. However, loose coupling limits performance in degraded scenarios where one sensor fails. 

Recent advancements focus on tightly-coupled schemes, typically fusing measurements within a factor graph \cite{feng2024block, shan2020lvi} or an Error-State Iterated Kalman Filter framework \cite{zheng2024fastlivo2, zheng2022fast, lin2021r2live, lin2022r3live}. LVI-SAM \cite{shan2020lvi} constructs a factor graph to fuse LiDAR-Inertial Odometry and Visual-Inertial Odometry, achieving high accuracy but with significant computational overhead due to global optimization. To improve efficiency, filter-based methods such as FAST-LIVO \cite{zheng2022fast}, FAST-LIVO2 \cite{zheng2024fastlivo2}, R2LIVE \cite{lin2021r2live} and R3LIVE \cite{lin2022r3live} integrate LiDAR, camera, and IMU states into a single ESIKF. More recently, AirSLAM \cite{xu2024airslam} and LIR-LIVO \cite{zhou2025lir} incorporates deep learning-based features to enhance robustness against illumination changes. Despite these successes, standard LIVO systems rely heavily on visible light cameras, rendering them fragile in the illumination-denied environments.

\subsection{Thermal Odometry}
Thermal infrared cameras offer a viable alternative for robust perception. Early works extended standard visual odometry algorithms to the thermal domain. Vidas et al. \cite{vidas2012hand} pioneered thermal odometry by employing optical flow to track GFTT features on images that had undergone rescaling. Shin et al. \cite{shin2019sparse} proposed a direct thermal-inertial SLAM method using sparse depth enhancement. However, thermal SLAM faces unique challenges compared to the visible spectrum. Thermal images typically suffer from low contrast, high noise, and lack of texture, which degrades the performance of handcrafted feature extractors \cite{Campos2021}. Another more critical issue is the AGC mechanism inherent in thermal cameras. As noted by Das et al. \cite{das2021online}, AGC induces drastic, non-linear radiometric rescaling to maximize contrast, violating the brightness constancy assumption and causing severe feature drift. Das et al. proposed a photometric calibration method to estimate gain and bias.

With the development of deep learning techniques, neural networks have been gradually introduced into the field of thermal odometry. Jiang et al. \cite{jiang2022thermal} combined traditional methods with deep learning to propose a real-time thermal odometry system: this system employs a Singular Value Decomposition-based image enhancement algorithm to improve feature detection performance and utilizes a lightweight optical flow network for feature tracking. Saputra et al. \cite{saputra2020deeptio} proposed DeepTIO, which constructs an end-to-end neural network architecture to realize pose regression-based thermal odometry. Nevertheless, standalone thermal imaging systems still struggle to meet the requirements for robustness and high precision in most practical scenarios. Meanwhile, we have observed that although thermal images suffer from weak texture and low contrast, they exhibit distinct structural features.

\subsection{Thermal Perception and Safety-Aware Mapping}
Thermal sensing enhances safety-aware perception, yet most SLAM systems decouple 2D thermal streams from 3D geometry, hindering automated hazard detection. We bridge this gap with a Probabilistic Intensity Voxel Map \cite{yuan2022efficient}. By leveraging high-precision pose estimates from our tightly coupled LiDAR-Inertial-Thermal framework, we anchor thermal measurements in a consistent global frame. This functional extension enables system detect the abnormal temperature variation alongside robust odometry.

\begin{figure}[!t] 
\setlength{\abovecaptionskip}{0.1cm} 
\includegraphics[width=\columnwidth]{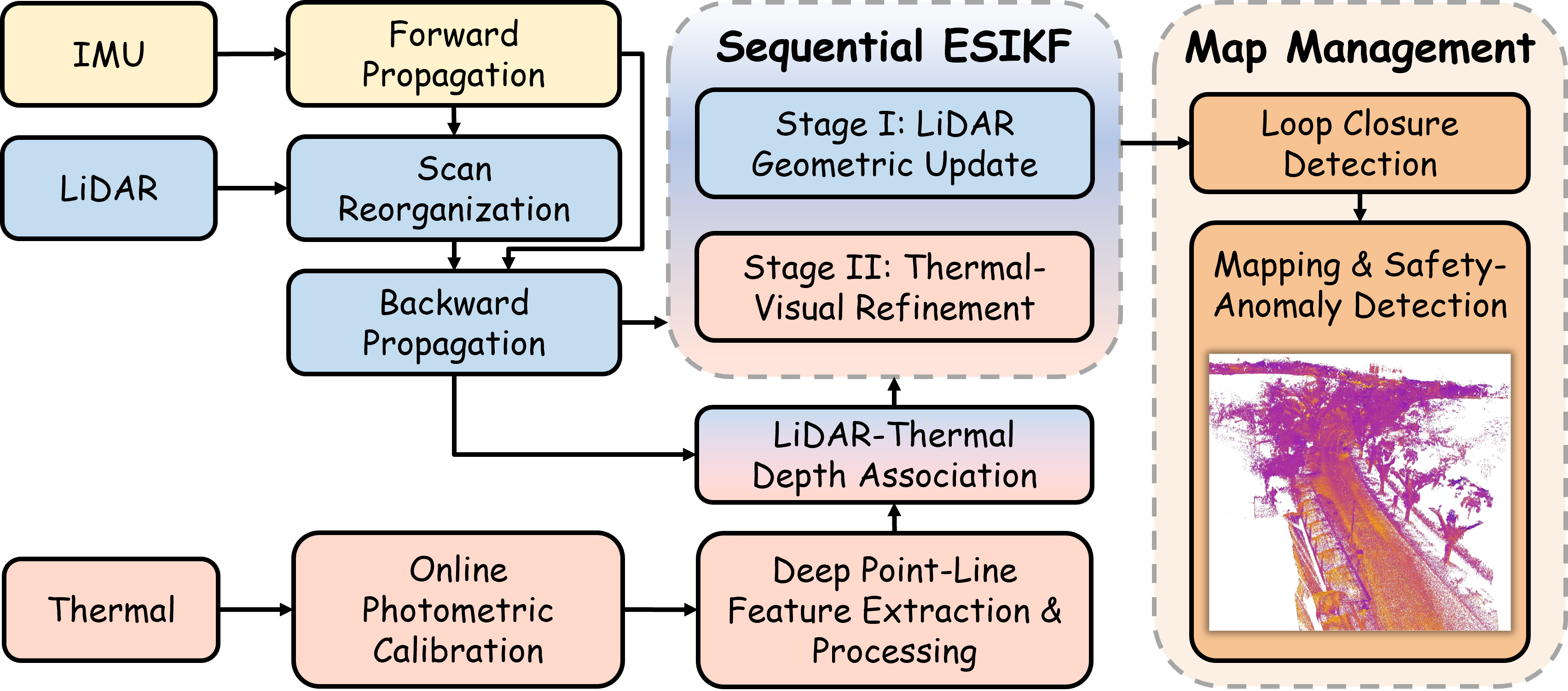}
\caption{The system overview of PL-LIT}
\label{fig: overview}
\vspace{-4mm}
\end{figure}

\section{System Overview}
The overall framework of our system is illustrated in Fig. \ref{fig: overview}. PL-LIT employs a Sequential ESIKF framework to sequentially fuse inertial, LiDAR, and thermal data. First, the LiDAR-Inertial module compensates for motion distortion and computes frame-to-map point-to-plane residuals to constrain the global geometric state. Subsequently, the thermal module refines the state using thermal-visual constraints. Specifically, we integrate an online photometric calibration module and utilize a fine-tuned PL-Net to extract point-line features, which are then associated with depth information provided by the local LiDAR map. A sliding window is employed to construct reprojection errors based on these point-line features. Finally, the optimized state is leveraged to project the calibrated thermal data into a Probabilistic Intensity Voxel Map. This system is also equipped with a loop closure detection function.


\section{State Estimation}


\begin{table}[t]
\caption{Mathematical Notations And Operators}
\label{tab:notations}
\centering
\begin{tabular}{p{3.0cm}|p{4.5cm}}
\toprule
\textbf{Notation} & \textbf{Description} \\
\midrule
${}^G(\cdot), {}^I(\cdot), {}^L(\cdot)$ & Global, IMU, and LiDAR frames. \\
${}^A\mathbf{T}_B = ({}^A\mathbf{R}_B, {}^A\mathbf{p}_B)$ & Transformation from frame $B$ to $A$. \\
$\boxplus, \boxminus$ & The Manifold boxplus and boxminus. \\
$\mathbf{x}, \hat{\mathbf{x}}, \tilde{\mathbf{x}}$ & True, predicted, and error state. \\
$[\mathbf{a}]_\wedge$ & Skew-symmetric matrix corresponding to vector $\mathbf{a} \in \mathbb{R}^3$. \\
\bottomrule
\end{tabular}
\end{table}

\subsection{State Definition and Forward Propagation}

The general mathematical notations are defined in Table \ref{tab:notations}. We define the system state vector $\mathbf{x}$ on the manifold $\mathcal{M}$ as:
\begin{equation}
    \mathbf{x} \triangleq [{}^G\mathbf{R}_I, {}^G\mathbf{p}_I, {}^G\mathbf{v}_I, \mathbf{b}_g, \mathbf{b}_a, {}^G\mathbf{g}]^\top
\end{equation}
where ${}^G\mathbf{R}_I \in SO(3)$ and ${}^G\mathbf{p}_I \in \mathbb{R}^3$ respectively denote the attitude and position of the IMU in the global frame. ${}^G\mathbf{v}_I$ is the IMU velocity, and ${}^G\mathbf{g}$ is the gravity vector. $\mathbf{b}_g$ and $\mathbf{b}_a$ represent the gyroscope and accelerometer biases, respectively.

To propagate the state forward, we discretize the continuous IMU kinematic model using a zero-order holder over the sampling interval $\Delta t$. The discrete-time state propagation is given by:
\begin{equation}
    \mathbf{x}_{i+1} = \mathbf{x}_{i} \boxplus \left( \Delta t \cdot \mathbf{f}(\mathbf{x}_{i}, \mathbf{u}_{i}, \mathbf{w}_i) \right)
\end{equation}
The state transition function $\mathbf{f}(\mathbf{x}, \mathbf{u}, \mathbf{w})$, driven by the IMU measurements $\mathbf{u} \triangleq [\boldsymbol{\omega}_m^\top, \mathbf{a}_m^\top]^\top$ and process noise $\mathbf{w} \triangleq [\mathbf{n}_g^\top, \mathbf{n}_a^\top, \mathbf{n}_{b_g}^\top, \mathbf{n}_{b_a}^\top]^\top$, are defined as:
\begin{equation}
\mathbf{f}(\mathbf{x}, \mathbf{u}, \mathbf{w}) = 
\begin{bmatrix}
\boldsymbol{\omega}_m - \mathbf{b}_g - \mathbf{n}_g \\
{}^G\mathbf{v}_I + \left( {}^G\mathbf{R}_I(\mathbf{a}_m - \mathbf{b}_a - \mathbf{n}_a) + {}^G\mathbf{g} \right) \frac{\Delta t}{2} \\
{}^G\mathbf{R}_I(\mathbf{a}_m - \mathbf{b}_a - \mathbf{n}_a) + {}^G\mathbf{g} \\
\mathbf{n}_{b_g} \\
\mathbf{n}_{b_a} \\
\mathbf{0}_{3 \times 1}
\end{bmatrix}
\end{equation}
The component $\mathbf{n}_g, \mathbf{n}_a$ represent the measurement noise, and $\mathbf{n}_{b_g}, \mathbf{n}_{b_a}$ represent the random walk noise for the biases.

We employ this model to predict the prior state, assuming zero noise ($\mathbf{w}=\mathbf{0}$):
\begin{equation}
    \hat{\mathbf{x}}_{i+1} = \hat{\mathbf{x}}_{i} \boxplus \left( \Delta t \cdot \mathbf{f}(\hat{\mathbf{x}}_{i}, \mathbf{u}_{i}, \mathbf{0}) \right)
\end{equation}
To facilitate linear error propagation, we introduce the error state $\widetilde{\mathbf{x}}$ as:
\begin{equation}
\begin{aligned}
    \widetilde{\mathbf{x}}_{i+1} &= \mathbf{x}_{i+1} \boxminus \hat{\mathbf{x}}_{i+1} \approx \mathbf{F}_{\widetilde{\mathbf{x}}} \widetilde{\mathbf{x}}_i + \mathbf{F}_{\mathbf{w}} \mathbf{w}_i
\end{aligned}
\end{equation}
Simultaneously, the error state covariance $\hat{\mathbf{P}}$ propagates as:
\begin{equation}
    \hat{\mathbf{P}}_{i+1} = \mathbf{F}_{\tilde{\mathbf{x}}}\hat{\mathbf{P}}_{i}\mathbf{F}_{\tilde{\mathbf{x}}}^\top + \mathbf{F}_{\mathbf{w}}\mathbf{Q}\mathbf{F}_{\mathbf{w}}^\top
\end{equation}
where $\mathbf{F}_{\tilde{\mathbf{x}}}$ and $\mathbf{F}_{\mathbf{w}}$ are the Jacobians with respect to the error state $\tilde{\mathbf{x}}$ and noise $\mathbf{w}$, respectively. $\mathbf{Q}$ is the covariance matrix of the process noise $\mathbf{w}$. This propagated state $\hat{\mathbf{x}}_{i+1}$ serves as the prior estimate for the subsequent measurement updates.

\subsection{LiDAR Measurement Model}


For a raw point ${}^L\mathbf{p}_j \in \mathcal{S}_k$, we first apply the Backward Propagation algorithm to compensate for motion distortion, projecting the point to the scan end time $t_k$. We then adopt a scan-to-map registration strategy. The corrected LiDAR point is transformed into the global frame, and a local plane is fit to its nearest neighbors in the global map.

Let ${}^G\mathbf{u}_j$ and ${}^G\mathbf{q}_j$ denote the normal vector and centroid of the corresponding planar patch. Considering the LiDAR measurement noise ${}^L\mathbf{n}_j$, the point-to-plane geometric residual $h_l(\cdot)$ is modeled as:
\begin{align}
0 &= h_l(\mathbf{x}_k, {}^L\mathbf{p}_j + {}^L\mathbf{n}_j) \nonumber \\
  &= {}^G\mathbf{u}_j^\top \left( {}^G\mathbf{R}_{I_k} \left( {}^I\mathbf{R}_L ({}^L\mathbf{p}_j + {}^L\mathbf{n}_j) + {}^I\mathbf{p}_L \right) + {}^G\mathbf{p}_{I_k} - {}^G\mathbf{q}_j \right)
\label{eq:lidar_residual}
\end{align}
The term ${}^I\mathbf{R}_L$ and ${}^I\mathbf{p}_L$ represent the extrinsic parameters between LiDAR and IMU. The formula calculates the perpendicular distance between the transformed LiDAR point and the map plane. Minimizing this residual converges the global geometric state, providing a high-accuracy linearization point for the subsequent thermal-visual refinement.

\subsection{Thermal-Visual Measurement Model}

This module refines the state by exploiting thermal visual constraints. Unlike standard visual-inertial systems, we address the unique challenges of thermal imaging—low contrast, drastic radiometric rescaling, lack of texture, and severe noise—through a structure-aware feature processing pipeline.

\subsubsection{Online Photometric Calibration}
While feature-based methods typically offer better robustness to photometric variations than direct methods, the AGC mechanism in thermal cameras introduces drastic, rapid, and nonlinear radiometric rescaling. This phenomenon severely disrupts inter-frame photometric consistency, leading to distribution shifts in feature descriptors and contrast degradation.

To address this radiometric inconsistency, we adopt an online parameter estimation approach based on Das et al. \cite{das2021online}. Prioritizing real-time system performance, we streamline the original method by omitting the computationally expensive spatial parameter estimation. Instead, we leverage the existing feature tracks from the visual frontend: the $(u, v)$ coordinates and intensity values are extracted directly from these tracks and utilized to estimate the global photometric parameters. We formulate this estimation as a nonlinear least squares problem with regularization:
\begin{equation}
    \min_{a, b} \left( \sum_{k \in \mathcal{K}} \rho \big( r_k(a, b) \big) + \lambda_1 \big((a - b) - 1\big)^2 + \lambda_2 b^2 \right)
\end{equation}
where $\mathcal{K}$ represents the set of valid feature correspondences. The photometric residual $r_k$ is defined by the mapping from the current frame to the reference keyframe:
\begin{equation}
    r_k(a, b) = I_{ref}(\mathbf{u}'_k) - \big[ (a - b) I_{curr}(\mathbf{u}_k) + b \big]
\end{equation}
In this formulation, $(a - b)$ and $b$ represent the effective gain and bias, respectively. The quadratic terms $\lambda_1 ((a - b) - 1)^2$ and $\lambda_2 b^2$ serve as regularization priors. Unlike penalizing the raw parameters directly, this decoupled formulation prevents parameter drift by constraining the effective gain $(a-b)$ towards unity and the bias $b$ towards zero, ensuring physical plausibility even under large parameter shifts. Finally, utilizing the estimated parameters, the calibrated intensity is computed as:
\begin{equation}
    I_{corr} = (a - b) I_{raw} + b
\end{equation}
where $\rho(\cdot)$ denotes a robust cost function employed to mitigate the influence of tracking outliers.

\subsubsection{LiDAR-Thermal Depth Association}
 The method associates high-precision LiDAR depth information with visual features to improve pose estimation accuracy. Within the camera view, LiDAR points maintained by a sliding window are projected onto a unit sphere. A 3D K-D tree finds some nearest points, which undergo a point-to-plane residual check to ensure validity.

For features with associated depth, a near-field focus area (critical for translation estimation and temperature anomaly detection) is divided into high-density subdivisions. Beyond this, points are divided into 10 uniform intervals, supplemented by extra feature extraction in sparse areas and near line features, ensuring even distribution.

\subsubsection{Deep Point-Line Feature Extraction}
To extract robust geometric features from blurred thermal images, we employ a fine-tuned PL-Net \cite{xu2024airslam} as the core extractor. Its backbone network follows the design philosophy of SuperPoint \cite{detone2018superpoint}. The line detection branch utilizes a stacked U-Net architecture to predict the Attraction Field representation, which encodes the distance and orientation parameters of potential lines. To efficiently bridge the domain gap between visible and thermal images, we adopt a transfer learning strategy: the pre-trained backbone and keypoint modules are frozen to retain generic geometric feature extraction capabilities, while the line detection module is fine-tuned on a small annotated thermal dataset with targeted augmentations. We converted the inference code from Python to C++, deployed it on NVIDIA TensorRT via ONNX.

\begin{figure}[!t]
\centering
\includegraphics[width=0.48\textwidth]{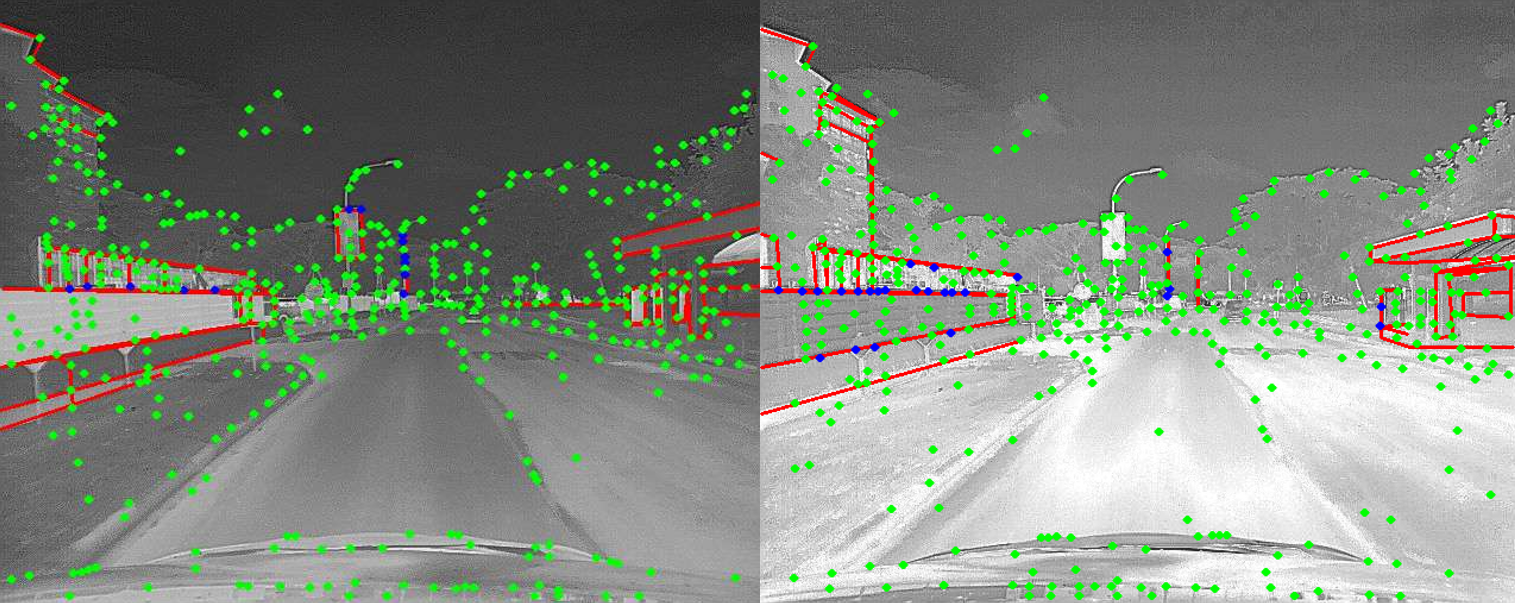}%
\caption{The front-end feature extraction effect of PL-LIT on a frame from the NTU4DRadLM \cite{lu2022ntu} loop3 sequence: the left image is the uncalibrated photometric image, and the right image is the photometrically calibrated one, where the blue points represent weighted points that satisfy the line feature. The right image has more blue points than the left image.}
\label{fig: feature}
\vspace{-4mm}
\end{figure}

\subsubsection{Point-Line Feature Processing}
Our visual module maintains a sliding window of keyframes. Depth information for feature points is directly associated from the synchronized LiDAR map, and feature matching adheres to the AirSLAM \cite{xu2024airslam} strategy. However, due to the thermal blurring effect, directly triangulating 3D lines often leads to severe bias.

Therefore, we propose a strategy that progresses from implicit weighting to explicit parameterization. For point features, structural information is exploited indirectly via an adaptive weighting scheme; however, for lines exhibiting long-term stability and redundant geometric support, we perform explicit 3D line parameterization to provide stronger geometric anchors. First, we employ a point-line association mechanism: for a pair of successfully matched feature points, we check whether both points are also successfully located on their corresponding matched line features in their respective image frames (the perpendicular distance from the point to the line is less than the threshold $\delta_{line}$). If this condition is met, the point pair is identified as Structure-Anchored Feature Points (e.g., the blue points shown in Fig. \ref{fig: feature}). These points are subsequently assigned a higher weight in the reprojection residual calculation, implicitly utilizing the structural information provided by the line features.

We formulate the visual measurement model based on the reprojection error of these depth-associated features. Let ${}^{C_k}\mathbf{p}_j = [u_j, v_j]^\top$ and ${}^{C_i}\mathbf{p}_j$ be the observed normalized coordinates of the $j$-th feature in the current frame $C_k$ and the reference frame $C_i$, respectively.

The 3D point in the reference frame, ${}^{C_i}\mathbf{P}_j$, is first recovered using its inverse depth $\delta_j = 1/d_j$ and inverse projection function $\pi^{-1}(\cdot)$:
\begin{equation}
    {}^{C_i}\mathbf{P}_j = \frac{1}{\delta_j} \pi^{-1}({}^{C_i}\mathbf{p}_j)
\end{equation}

This point is then transformed to the current camera frame $C_k$ via the kinematic chain involving the IMU states and extrinsic parameters:
\begin{equation}
\begin{aligned}
    {}^{C_k}\mathbf{P}_j &= {}^I\mathbf{R}_C^\top \bigg( {}^G\mathbf{R}_{I_k}^\top \Big( {}^G\mathbf{R}_{I_i} ({}^I\mathbf{R}_C {}^{C_i}\mathbf{P}_j + {}^I\mathbf{p}_C) \\
    &\quad + {}^G\mathbf{p}_{I_i} - {}^G\mathbf{p}_{I_k} \Big) - {}^I\mathbf{p}_C \bigg)
\end{aligned}
\end{equation}
where ${}^I\mathbf{R}_C$ and ${}^I\mathbf{p}_C$ denote the extrinsic rotation and position from the Camera frame to the IMU frame.

The reprojection residual $r_{\mathcal{C}}$ is defined as the difference between the observation and the projection of the predicted 3D point ${}^{C_k}\mathbf{P}_j$:
\begin{equation}
    r_{\mathcal{C}}(\mathbf{x}_k, {}^{C_k}\mathbf{p}_j) = {}^C\mathbf{p}_j - \pi( {}^{C_k}\mathbf{P}_j )
\end{equation}
Here, $\pi(\cdot)$ is the projection function. To incorporate structural information from line features, we assign higher weights to points that lie close to detected line segments using a Gaussian weighting function. We design the adaptive confidence factor $w_j$ as:
\begin{equation}
    w_j = 
    \begin{cases} 
    \eta \cdot (1 + e^{-\frac{dist(j, l)^2}{\sigma^2}}) & \text{if } j \in \text{Structure-Anchored} \\
    1 & \text{otherwise}
    \end{cases}
\end{equation}
where $dist(j, l)$ is the perpendicular distance between point $j$ and its associated line $l$. $\eta > 1$ is a gain coefficient and $\sigma$ controls the sensitivity. 

To mathematically incorporate this structural confidence into the optimization, the measurement information matrix for the $j$-th visual feature is scaled as $\mathbf{R}_{\mathcal{C}, j}^{-1} = w_j^2 \mathbf{R}_0^{-1}$, with $\mathbf{R}_0$ being the nominal noise covariance. In the context of the nonlinear least-squares problem, this is equivalent to scaling the Mahalanobis distance of the reprojection residual by $w_j$. This adaptive weighting scheme ensures that features tightly aligned with structural lines correspond to higher stiffness in the normal equations, thereby effectively utilizing the constraints introduced by successfully matched line features in thermal images.

\subsubsection{Spatiotemporal Explicit Structural Constraints}
While the implicit weighting strategy effectively handles local features, thermal imagery often provides distinct global edges (e.g., corridors, doors) that remain stable even when texture is absent. To fully leverage these global geometric primitives, we upgrade valid implicit constraints to explicit 3D line constraints using a spatiotemporal filter.

A 2D line segment is promoted to an explicit constraint if it satisfies two strict criteria: 1) Temporal Consistency: The line is successfully tracked for at least $N_{track}$ consecutive frames; 2) Spatial Redundancy: To overcome the sparsity of single-frame LiDAR measurements, we aggregate the associated line-depth points from all tracked frames into a accumulated point set $\mathcal{U}_L$ by transforming them using their respective estimated poses.

For such candidates, we employ Principal Component Analysis on the accumulated set $\mathcal{U}_L$ to robustly initialize the 3D line via Total Least Squares fitting. This yields the line's centroid $\boldsymbol{\mu}_L$ and principal direction vector $\mathbf{v}_L$. The 3D line $L_w$ in the world frame is then parameterized using Plücker coordinates $\mathcal{L}_w = (\mathbf{n}_w, \mathbf{d}_w)^\top$, where the direction vector $\mathbf{d}_w = \mathbf{v}_L$ and the moment vector $\mathbf{n}_w = \boldsymbol{\mu}_L \times \mathbf{d}_w$. Therefore, using the above-described method, we leverage the LiDAR-enhanced depth correlation mechanism to directly obtain high-precision 3D coordinates of the upsampled points on line features. Subsequently, we can robustly recover the geometric parameters of the 3D lines, thereby avoiding the scale ambiguity and initialization instability caused by monocular camera-based triangulation in thermal imaging SLAM.

The line is transformed to the current camera frame $C_k$ as $\mathcal{L}_{c_k} = (\mathbf{n}_{c_k}, \mathbf{d}_{c_k})^\top$ via the line transformation derived from the camera pose. The projection of this 3D line onto the image plane is determined by the moment vector $\mathbf{n}_{c_k}$, which represents the normal of the interpretation plane. The reprojection residual $r_{\mathcal{L}}$ is formulated as the point-to-line distance between the observed 2D endpoints ($\mathbf{s}, \mathbf{e}$) and the projected line normal $\mathbf{l}_{proj} = [l_1, l_2, l_3]^\top = \mathbf{n}_{c_k}$:
\begin{equation}
    r_{\mathcal{L}}(\mathbf{x}_k, \mathcal{L}_w) = \begin{bmatrix} 
    \frac{\mathbf{s}^\top \mathbf{l}_{proj}}{\sqrt{l_1^2 + l_2^2}} \\
    \frac{\mathbf{e}^\top \mathbf{l}_{proj}}{\sqrt{l_1^2 + l_2^2}}
    \end{bmatrix}
\end{equation}
This explicit constraint allows feature points to slide along the edge, strictly penalizing only the perpendicular geometric error. The feature points that are successfully utilized to initialize and constrain explicit 3D lines are excluded from the point-based visual reprojection term ($r_{\mathcal{C}}$).

\subsection{Sequential ESIKF}
To efficiently fuse the heterogeneous multi-modal measurements while maintaining real-time performance, we employ a two-stage sequential update strategy similar to FAST-LIVO2 \cite{zheng2024fastlivo2} within a tightly-coupled ESIKF framework: \ding{172} LiDAR Geometric Update; \ding{173} Thermal-Visual Refinement.

\subsubsection{Stage I: LiDAR Geometric Update}
In the first stage, we utilize the IMU-propagated state $\hat{\mathbf{x}}_k$ and covariance $\hat{\mathbf{P}}_k$ (from Section IV-A) as the prior. The optimal error state $\tilde{\mathbf{x}}$ is estimated by minimizing the LiDAR geometric residual $h_l(\cdot)$ defined in Eq.\eqref{eq:lidar_residual}. This is formulated as an Iterated EKF (IESKF) optimization problem:
\begin{equation}
    \min_{\tilde{\mathbf{x}}} \left( \| \tilde{\mathbf{x}} \|_{\hat{\mathbf{P}}_k^{-1}}^2 + \sum_{j \in \mathcal{S}_k} \| h_l(\hat{\mathbf{x}}_k \boxplus \tilde{\mathbf{x}}, {}^L\mathbf{p}_j + {}^L\mathbf{n}_j) \|_{\mathbf{R}_{\mathcal{L}}^{-1}}^2 \right)
\end{equation}
where $\mathbf{R}_{\mathcal{L}}$ is the covariance matrix of LiDAR measurement noise. We iteratively solve this optimization. In each iteration, the Kalman gain $\mathbf{K}_{\mathcal{L}}$ and the state update are computed. This intermediate posterior state, denoted as $\hat{\mathbf{x}}_{k}^{\mathcal{L}}$, provides a geometrically consistent linearization point for the subsequent visual module.

\subsubsection{Stage II: Thermal-Visual Refinement}
In the second stage, we refine the state using both structure-anchored feature points and explicit line constraints. The intermediate state $\hat{\mathbf{x}}_{k}^{\mathcal{L}}$ and its covariance $\mathbf{P}^{\mathcal{L}}$ serve as the prior. The optimization objective incorporates the weighted visual reprojection residuals $r_{\mathcal{C}}$ and the explicit line residuals $r_{\mathcal{L}}$:
\begin{equation}
    \min_{\tilde{\mathbf{x}}} \left( \| \tilde{\mathbf{x}} \|_{(\mathbf{P}^{\mathcal{L}})^{-1}}^2 + \sum_{j \in \mathcal{F}_k} \| r_{\mathcal{C}} \|_{\mathbf{R}_{\mathcal{C}, j}^{-1}}^2 + \sum_{m \in \mathcal{E}_k} \| r_{\mathcal{L}} \|_{\mathbf{R}_{\mathcal{E}}^{-1}}^2 \right)
\end{equation}
Here, $\mathcal{F}_k$ represents the set of point features, and $\mathcal{E}_k$ denotes the set of explicit line constraints verified by our spatiotemporal filter. This dual-constraint formulation ensures that the system benefits from the texture-based details of point features while maintaining global geometric consistency via explicit lines, effectively dominating the gradient descent direction in degenerate scenarios. The final posterior state $\hat{\mathbf{x}}_{k+1}$ and covariance $\hat{\mathbf{P}}_{k+1}$ are obtained after this refinement, completing the tightly-coupled fusion process.

\section{Map Management}
\subsection{Loop Closure Detection}
Because of the thermal images, traditional visual Bag-of-Words loop closure schemes struggle due to poor feature robustness and frequent mismatches. So our solution integrates a geometry-centric loop closure module \cite{wang_fast_lio_lc}. This approach identifies loop candidates via a spatiotemporal radius search. Potential candidates are then strictly verified and finely registered against the current frame using Iterative Closest Point. Constraints with a high geometric fitness score are accepted and optimized within a GTSAM factor graph.

\subsection{Mapping and Safety-Anomaly Detection}

 We note that the key to most heat-source hazard problems lies not in the absolute temperature threshold, but in whether temperature changes can be detected to exceed the threshold. Standard 8-bit data streams are computationally efficient, but are susceptible to drift induced by Automatic Gain Control. To address these issues, the PL-LIT algorithm employs a lightweight probabilistic intensity mapping strategy that decouples temperature information maintenance from visualization functions.

We utilize the intensities output by the Online Photometric Calibration module to update the statistical model of the global Probabilistic Voxel Map. Each voxel maintains a self-adaptive Gaussian distribution $\mathcal{N}(\mu_I, \sigma_I^2)$ based on calibrated data, enabling robust detection of thermal anomalies against AGC drift. Simultaneously, we retain the raw 8-bit grayscale values for point cloud colorization. This design facilitates secondary human verification in the event of a safety alarm and enables high-contrast pseudo-color map visualization.

To achieve robust identification of hazards, we design a multi-stage detection mechanism, which includes setting a distance threshold between the detection area and the robot, defining the \(3\sigma\) criterion for grayscale variations, and establishing a minimum absolute threshold, among other measures.

\section{Experiments}

To comprehensively evaluate the performance of PL-LIT, we design three groups of experiments.

First, we utilize the Hilti Challenge 2022 dataset \cite{9968057}, which provides high-precision ground truth and RGB camera measurements, to verify the generality and robustness of PL-LIT in LiDAR–Inertial–Visual configurations under partially degraded visible-light conditions. The evaluation is conducted across three difficulty levels: easy (e.g., exp04, exp05; typical motion and scene complexity), medium (e.g., exp06, exp14; aggressive and rapid motions), and hard (e.g., exp16, exp18; long corridors, weak textures with aggressive and rapid motions). This hierarchical evaluation enables a systematic assessment of the system’s robustness against geometric degeneration, illumination variation, and motion-induced challenges.

Second, we employ the NTU4DRadLM dataset \cite{lu2022ntu} to validate the accuracy and robustness of PL-LIT on long-range thermal sequences. This dataset primarily consists of semi-structured urban road environments characterized by abundant line features and large-scale trajectories, providing a suitable benchmark to highlight the advantages of our structure-aware point–line fusion framework in thermal imaging scenarios.

Finally, to demonstrate the practical engineering value of the proposed system, we construct a handheld hardware platform (as shown in Fig. \ref{fig:Safety}(d)) and conduct anomaly detection mapping experiments. The handheld device is equipped with a Livox Horizon LiDAR, a FLIR RGB camera (BFS-U3-13Y3), a Botu thermal camera (TB640-4M), and a Hipnuc IMU (HI71T3-MI0). These experiments evaluate the real-world applicability of PL-LIT for thermal anomaly detection and safety inspection. Since few publicly available SLAM systems support 
thermal imagery, we compare PL-LIT against state-of-the-art LiDAR–Inertial–Visual Odometry (LIVO) and LiDAR–Inertial Odometry (LIO) frameworks, including FAST-LIVO2 \cite{zheng2024fastlivo2}, FAST-LIVO \cite{zheng2022fast}, R3LIVE \cite{lin2022r3live}, and FAST-LIO2 \cite{Xu2022}, using default or identical parameter settings whenever applicable. For a fair comparison, the loop closure module in PL-LIT is disabled in all benchmarking experiments. Accuracy is quantified using the Root Mean Square Error (RMSE) of the Absolute Pose Error (APE).

\begin{figure}[!t]
\centering
\includegraphics[width=0.48\textwidth]{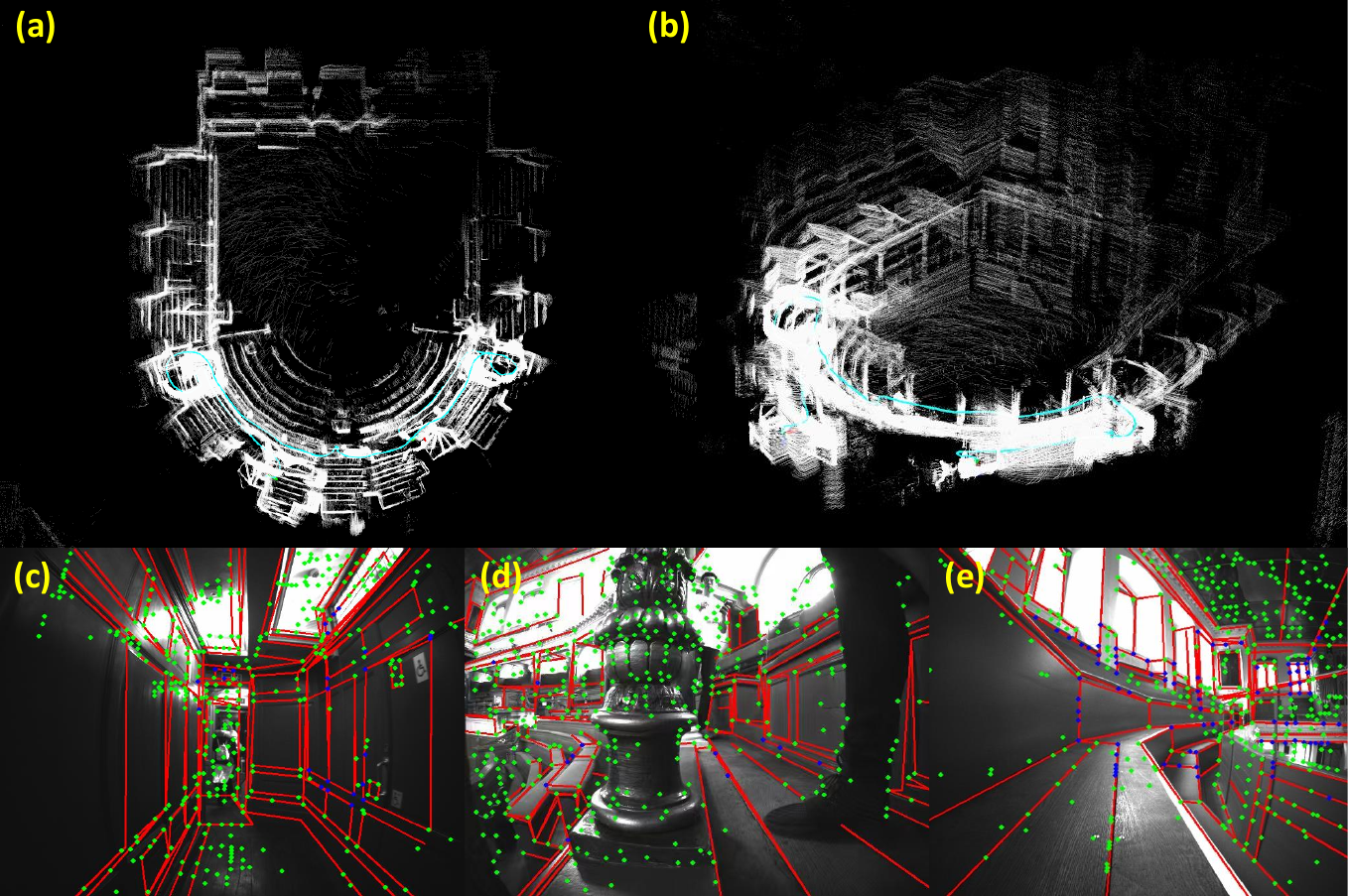}%
\caption{(a)(b) present the mapping performance of PL-LIT in the hard sequence exp18 of the Hilti dataset. (c)(d)(e) illustrate the front-end point-line feature extraction results.}
\label{fig: LIV_figure}
\end{figure}

\begin{figure}[!t]
\centering
\includegraphics[width=0.48\textwidth]{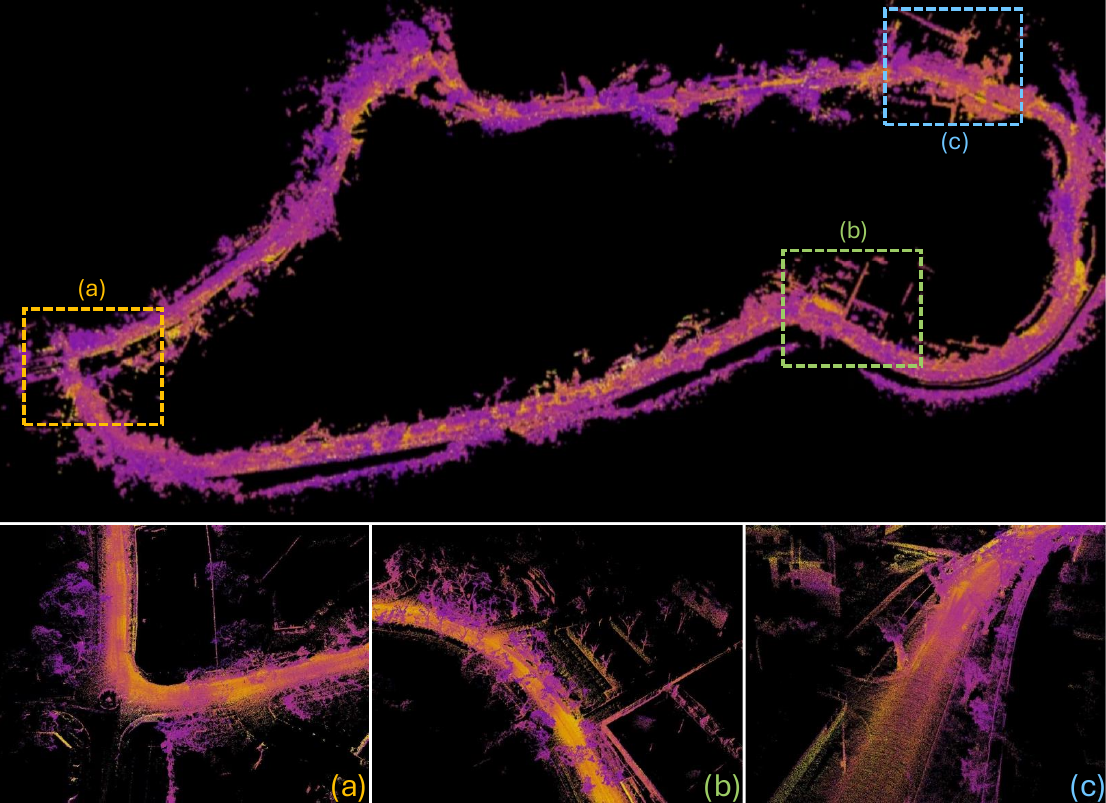}%
\caption{This figure presents the radiometric mapping performance in the NTU4DRadLM loop3 sequence. The bottom row shows zoomed-in views of the highlighted regions in (a), (b), and (c), respectively.}
\label{fig: LIT_figure}
\vspace{-4mm}
\end{figure}

\begin{table*}[!t]
\caption{Comparison of RMSE APE [m] in Visible-Light and Thermal Environments}
\label{tab:rmse_ate_combined}
\centering
\setlength{\tabcolsep}{5pt}

\begin{tabularx}{\textwidth}{l|*{6}{>{\centering\arraybackslash}X}|*{3}{>{\centering\arraybackslash}X}}
\toprule
\multirow{3}{*}{\textbf{Methods}}
& \multicolumn{6}{c|}{\textbf{Hilti Challenge 2022} {\footnotesize\textcolor{gray}{(LiDAR-Inertial-Visual)}}}
& \multicolumn{3}{c}{\textbf{NTU4DRadLM} {\footnotesize\textcolor{gray}{(LiDAR-Inertial-Thermal)}}} \\
\cmidrule(lr){2-7} \cmidrule(lr){8-10}
 & \textbf{exp04} & \textbf{exp05} & \textbf{exp06} & \textbf{exp14} & \textbf{exp16} & \textbf{exp18}
 & \textbf{loop1} & \textbf{loop2} & \textbf{loop3} \\
 & \footnotesize{Easy} & \footnotesize{Easy} & \footnotesize{Medium}
 & \footnotesize{Medium} & \footnotesize{Hard} & \footnotesize{Hard}
 & \footnotesize{(6.95km)} & \footnotesize{(4.79km)} & \footnotesize{(4.23km)} \\
\midrule

FAST-LIO2 \cite{Xu2022}
& \cellcolor{secondcolor}0.055 & 0.058 & 0.047 & 0.094 & $\times$ & 0.614
& \cellcolor{secondcolor}11.378 & \cellcolor{secondcolor}11.376 & \cellcolor{secondcolor}12.289 \\

R3LIVE \cite{lin2022r3live}
& 0.197 & 0.181 & 0.098 & 0.277 & $\times$ & $\times$
& 14.735 & 13.639 & 13.045 \\

FAST-LIVO \cite{zheng2022fast}
& 0.210 & 0.152 & 0.102 & 0.224 & $\times$ & $\times$
& 22.419 & 12.496 & 13.790 \\

FAST-LIVO2 \cite{zheng2024fastlivo2}
& \cellcolor{bestcolor}0.049 & \cellcolor{bestcolor}0.044 & \cellcolor{bestcolor}0.039
& \cellcolor{bestcolor}0.037 & \cellcolor{bestcolor}0.373 & \cellcolor{bestcolor}0.242
& 11.450 & 11.885 & 12.691 \\

\textbf{PL-LIT (Ours)}
& 0.056 & \cellcolor{secondcolor}0.052 & \cellcolor{secondcolor}0.045
& \cellcolor{secondcolor}0.058 & \cellcolor{secondcolor}0.943 & \cellcolor{secondcolor}0.372
& \cellcolor{bestcolor}7.651 & \cellcolor{bestcolor}8.301 & \cellcolor{bestcolor}8.442 \\

\bottomrule
\end{tabularx}

\vspace{-2pt}
\begin{flushleft}
\footnotesize Cells highlighted in \bestbox{} and \secondbox{} indicate the best and the second-best performance, respectively.
$\times$ denotes the system totally failed.
\end{flushleft}
\end{table*}

\subsection{Performance on LiDAR-Inertial-Visual Dataset}

Although PL-LIT is primarily designed as a LiDAR–Inertial–Thermal fusion framework, our algorithm is designed to address the weak texture problem in thermal images. Therefore, it demonstrates highly competitive performance on difficult LiDAR–Inertial–Visual datasets as well. As shown in Table \ref{tab:rmse_ate_combined} and Fig. \ref{fig: LIV_figure}, under both nominal conditions (exp04–exp14), PL-LIT consistently outperforms FAST-LIO2, R3LIVE, and FAST-LIVO.

In the most challenging sequences (exp16 and exp18), characterized by long corridors, weak textures, aggressive motion, and severe illumination variations, LiDAR-only odometry almost fails tracking, while conventional visual-inertial approaches suffer from violations of the brightness constancy assumption, leading to system divergence. In contrast, by incorporating structure-aware point–line features, PL-LIT significantly improves both robustness and accuracy.

Although PL-LIT yields slightly higher errors than FAST-LIVO2, the latter incorporates substantial refinements in both LiDAR odometry and visual preprocessing to enhance robustness against geometric and photometric degradation. In contrast, PL-LIT retains the original LiDAR odometry module of FAST-LIVO without modification. Nevertheless, incorporating structure-aware features in a tightly coupled LiDAR–Inertial-Visual framework yields consistent performance gains.

Notably, although PL-LIT is primarily designed to address weak-texture challenges in thermal imaging, it maintains high accuracy in challenging visible-light sequences with similar texture deficiencies. This demonstrates that the proposed structure-aware formulation is not modality-specific, but generalizes effectively across both visible and thermal sensing conditions.

\subsection{Performance on LiDAR-Inertial-Thermal Dataset}

We further evaluate the proposed framework on real thermal data using the loop 1 to 3 sequences from the NTU4DRadLM dataset. These sequences consist of long-range outdoor trajectories (over 4 km) and present substantial challenges due to thermal sensor noise, low contrast, and AGC-induced radiometric inconsistency.

\begin{figure}[!t]
\centering
\hspace{-0.5cm} 
\includegraphics[width=0.45\textwidth]{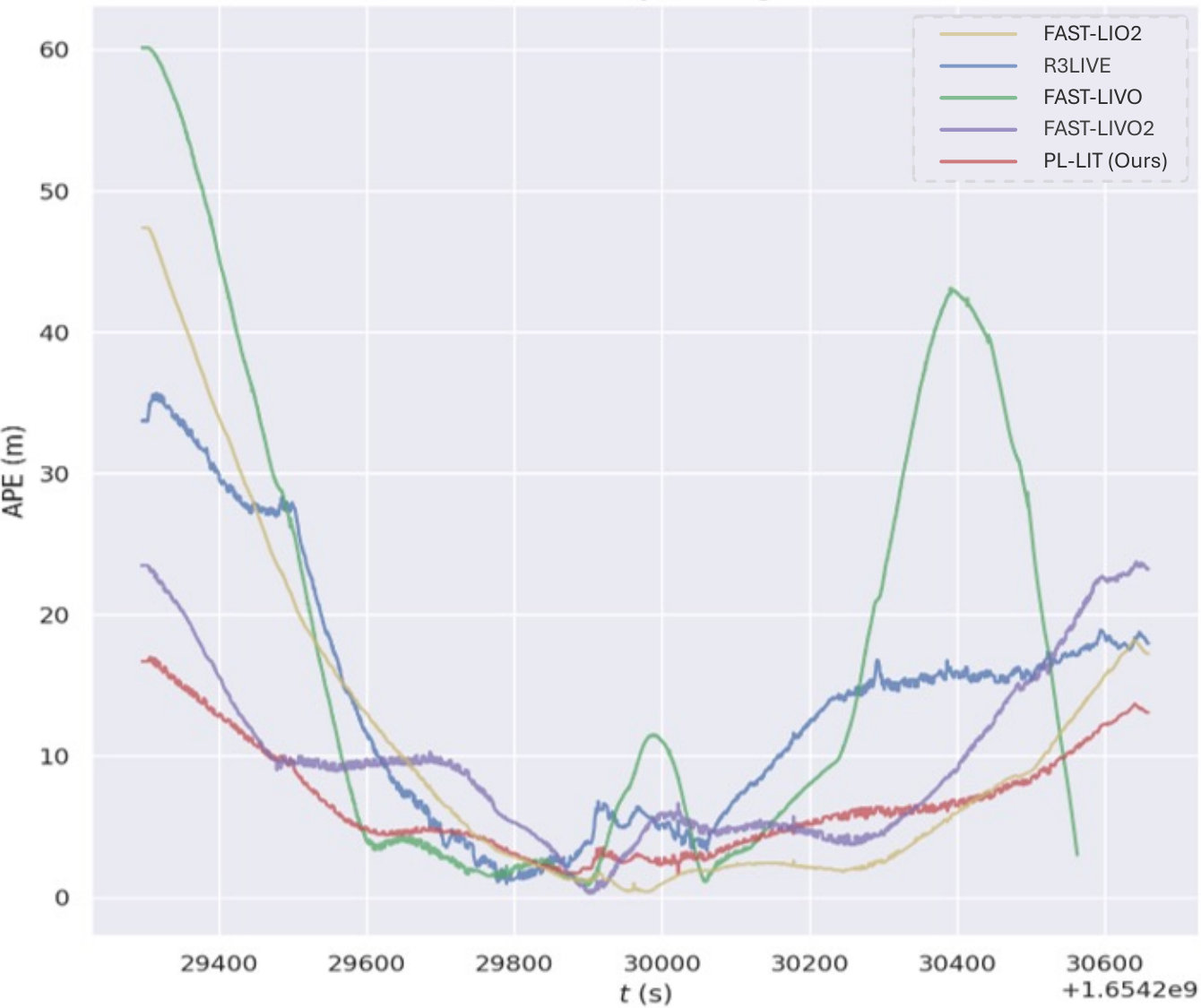}%
\caption{APE comparison in the NTU4DRadLM loop1 sequence.}
\label{fig:evo}
\vspace{-4mm}
\end{figure}

The quantitative results are summarized in Table \ref{tab:rmse_ate_combined} and Fig. \ref{fig:evo}. Notably, conventional LIVO methods exhibit increased trajectory errors in these thermal scenarios, even performing 
worse than pure LIO framework (FAST-LIO2). This behavior suggests that raw thermal imagery, when processed using standard visual pipelines, may introduce unstable photometric constraints due to weak texture and radiometric distortion, thereby limiting its contribution to state estimation.

In contrast, PL-LIT achieves the best performance across all thermal sequences, reducing the RMSE by 27-33\% compared to FAST-LIVO2 and FAST-LIO2. This improvement demonstrates that the proposed online photometric calibration and structure-aware point–line modeling effectively extract stable geometric constraints from thermal imagery, transforming it from a noisy modality into a reliable source of complementary information.
Fig. \ref{fig: LIT_figure} and Fig. \ref{fig:evo} further illustrates the qualitative performance.

\subsection{Thermographic Mapping and Anomaly Detection}

Beyond evaluating state estimation accuracy, we further demonstrate the system’s applicability to safety inspection tasks. Using our handheld platform, we construct a Probabilistic Intensity Voxel Map and evaluate the real-time thermal anomaly detection capability.

During the experiment, the system repeatedly traversed an indoor environment in cyclic motion. In the initial pass, all objects on a table exhibited temperatures consistent with the ambient environment, allowing the system to establish a baseline probabilistic intensity model. Before completing the first cycle and revisiting the table area, a heated object was introduced to simulate a sudden thermal anomaly (e.g., equipment overheating).

Upon re-observation, the system detected the anomaly as a statistically significant deviation from the previously estimated voxel intensity distribution. The anomalous regions were marked in real time using red point clouds, and the voxel visualization was dynamically adjusted to enhance interpretability.

\begin{figure}[!t]
\centering
\includegraphics[width=0.48\textwidth]{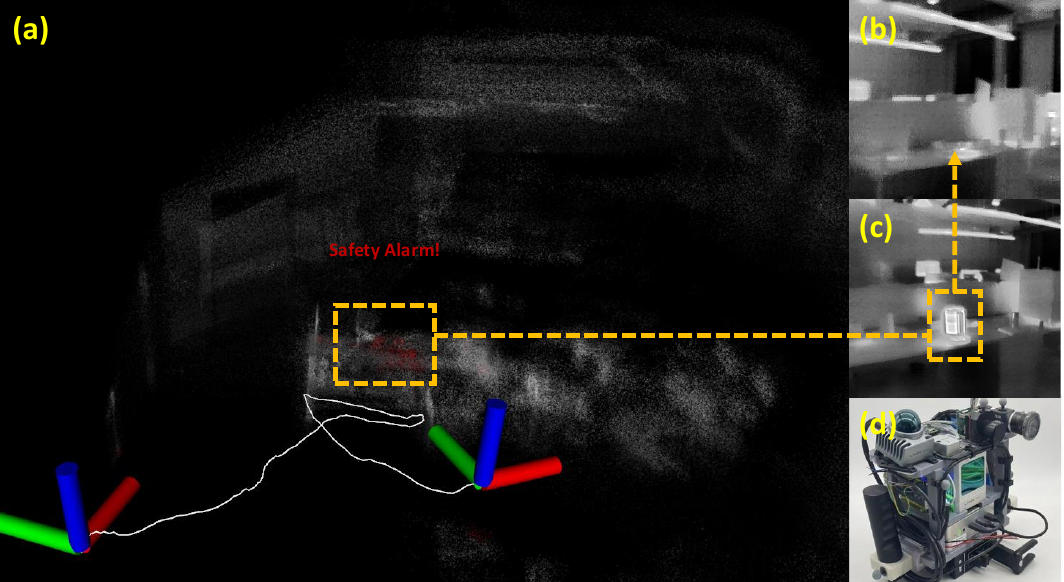}%
\caption{The figure shows the visualization of anomaly detection and safety alarm in the experiment. (b) and (c) are the thermal images taken before and after placing a boiled kettle. (d) is our self-designed handheld device.}
\label{fig:Safety}
\vspace{-4mm}
\end{figure}

The experiment was repeated three times with varying anomaly locations, achieving a 100\% success rate. However, the spatial accuracy of the estimated anomaly positions remains limited. In addition, the detection of anomalous heat values is only applicable over relatively short periods and cannot operate accurately around the clock. These issues will be further optimized in future work. Fig. \ref{fig:Safety} illustrates the detection and alarm process.

\section{Conclusions and Future Works}  
We presented PL-LIT, a robust LiDAR-Inertial-Thermal SLAM system. By integrating online photometric calibration and a deep-learning-based point-line feature extraction network, the system effectively overcomes the challenges of photometric instability and texture sparsity in thermal imagery. The tightly coupled sequential ESIKF framework ensures high-precision state estimation in visible-light environments and long-range thermal environments. Furthermore, the construction of a Probabilistic Intensity Voxel Map enables real-time thermal anomaly monitoring as a functional extension of the odometry pipeline. Evaluations demonstrate that PL-LIT achieves superior accuracy and robustness compared to existing frameworks.

In future work, we plan to enhance the loop closure module by integrating deep thermal descriptors with geometric constraints, further improving system stability in geometrically degenerate environments such as long tunnels or open fields. Additionally, we will investigate a tighter integration of photometric calibration parameters into the global optimization backend to ensure stable and consistent thermographic mapping under drastic ambient temperature variations. Finally, we aim to incorporate more robust LiDAR odometry frameworks to further improve overall system accuracy and robustness.

\bibliographystyle{IEEEtran}        
\bibliography{ref}

\end{document}